\documentclass[lettersize,journal]{IEEEtran}
\usepackage{amsmath}
\usepackage{threeparttable}
\usepackage[utf8]{inputenc}
\usepackage[english]{babel}
\usepackage{booktabs} 
\usepackage[absolute,overlay]{textpos}
\usepackage{etoolbox}
\usepackage{subfigure}
\usepackage[export]{adjustbox}
\usepackage[linesnumbered,ruled,vlined]{algorithm2e}
\usepackage{color, colortbl}
\usepackage{longtable}
\usepackage{xcolor}
\usepackage{soul}
\usepackage{comment}
\usepackage{url}
\usepackage{balance}
\usepackage{graphicx}
\usepackage{multirow}
\usepackage{textcomp}
\usepackage{lipsum}
\usepackage{tikz}
\usepackage{hyperref}
\usepackage{cleveref}
\usepackage{float}
\usepackage{xspace}
\usepackage{stfloats}
\usepackage{hhline}
\usepackage{enumitem}
\newbool{showComments}
\booltrue{showComments}

\ifbool{showComments}{%

}

\newcommand{\sysname}{EmbodiTTA\xspace}

\ifCLASSINFOpdf
\else
\fi
\begin{document}
%
\title{EmbodiTTA: Resource-Efficient Test-Time Adaptation for Embodied Visual Systems}
%
%
%

\author{Xiao Ma,
        Young~D.~Kwon,~\IEEEmembership{Member,~IEEE,}
        and~Dong~Ma\thanks{*Dong Ma is the corresponding author.},~\IEEEmembership{Member,~IEEE}
\thanks{Xiao Ma is with the Singapore Management University, Singapore,
e-mail: xiaoma.2022@phdcs.smu.edu.sg.}
\thanks{Young~D.~Kwon is with Samsung AI Center--Cambridge, United Kingdom, e-mail: yd.kwon@samsung.com}
\thanks{Dong Ma is with University of Cambridge, United Kingdom, e-mail: dm878@cam.ac.uk.}
\thanks{Copyright (c) 20xx IEEE. Personal use of this material is permitted. However, permission to use this material for any other purposes must be obtained from the IEEE by sending a request to pubs-permissions@ieee.org.}
}

%
%

\markboth{IEEE Internet of Things Journal}%
{EmbodiTTA: Resource-Efficient Test-Time Adaptation for Embodied Visual Systems}
%



\maketitle


\begin{abstract}

Deep neural networks (DNNs) are widely used for perception in embodied visual systems such as drones and AR glasses, where models must process streaming data under strict memory and energy budgets. However, once deployed in the physical world, these models often suffer from performance degradation in dynamic scenarios due to domain shifts. Although previous research has proposed test-time adaptation (TTA) to address this issue, which continuously adapts the deployed model \textit{on every incoming batch of data before performing actual inference}, these approaches are impractical and inefficient from a system perspective, as they incur substantial memory and energy consumption and introduce high latency. In this work, we rethink the conventional paradigm of existing TTA and introduce a novel paradigm -- on-demand TTA -- which triggers adaptation only when a significant domain shift is detected.
Then, we present \sysname, an efficient on-demand TTA framework for \textit{accurate and efficient} TTA on embodied devices. \sysname comprises three techniques: 1) a lightweight domain shift detection mechanism to activate on-demand TTA only when needed, drastically reducing the overall computational overhead, 2) a source domain selection module to choose the appropriate starting point for adaptation, and 3) a decoupled Batch Normalization (BN) update scheme to reduce the memory overhead and achieve stable adaptation. 
We evaluated \sysname on the Jetson Orin Nano for both object recognition and semantic segmentation tasks. The results show that \sysname achieves the highest TTA accuracy across nearly all tasks and batch size settings—and is uniquely the only BN-based method effective at batch size 1—while reducing latency by up to 6.7$\times$ and energy consumption by 47.2\% compared to the baselines, with comparable peak memory usage. Code is released at \url{https://github.com/TheMaXiao/EmbodiTTA}.


\end{abstract}
\begin{IEEEkeywords}
Embodied AI, Mobile Vision, Domain Adaptation, Test-Time Adaptation
\end{IEEEkeywords}

\ifCLASSOPTIONpeerreview
\begin{center} \bfseries EDICS Category: 3-BBND \end{center}
\fi

\IEEEpeerreviewmaketitle

\section{Introduction}


Deep neural networks (DNNs) are becoming a core component of intelligent Internet-of-Things (IoT) and embodied AI systems~\cite{liu2025embodied}, enabling egocentric (first-person) perception and on-device intelligence in resource-constrained platforms such as mobile robots, drones and AR/VR glasses, where models must process live sensory streams in real time. In these IoT-enabled embodied systems, visual perception often serves as the first step, providing scene understanding for downstream decision-making and interaction. However, as a data-driven technique, DNNs typically \textit{achieve optimal performance only when training and testing data share the same distribution}~\cite{geirhos2018generalisation,recht2019imagenet}. In the physical world, embodied systems inevitably encounter dynamic and unseen data distributions in their streaming test inputs, a phenomenon known as \textit{domain shift}, caused by factors such as weather changes, sensor noise, and varying lighting conditions, which can lead to substantial performance degradation~\cite{hendrycks2019benchmarking}. For example, a battery-powered delivery robot (or drone) must recognize obstacles in real time on an edge accelerator; when it moves from indoor to outdoor or from daylight to night, the same objects can look very different, causing recognition errors while the device still has to meet tight latency, memory, and energy budgets.

\begin{figure}[t]
\centering
  \includegraphics[width = 0.45\textwidth]{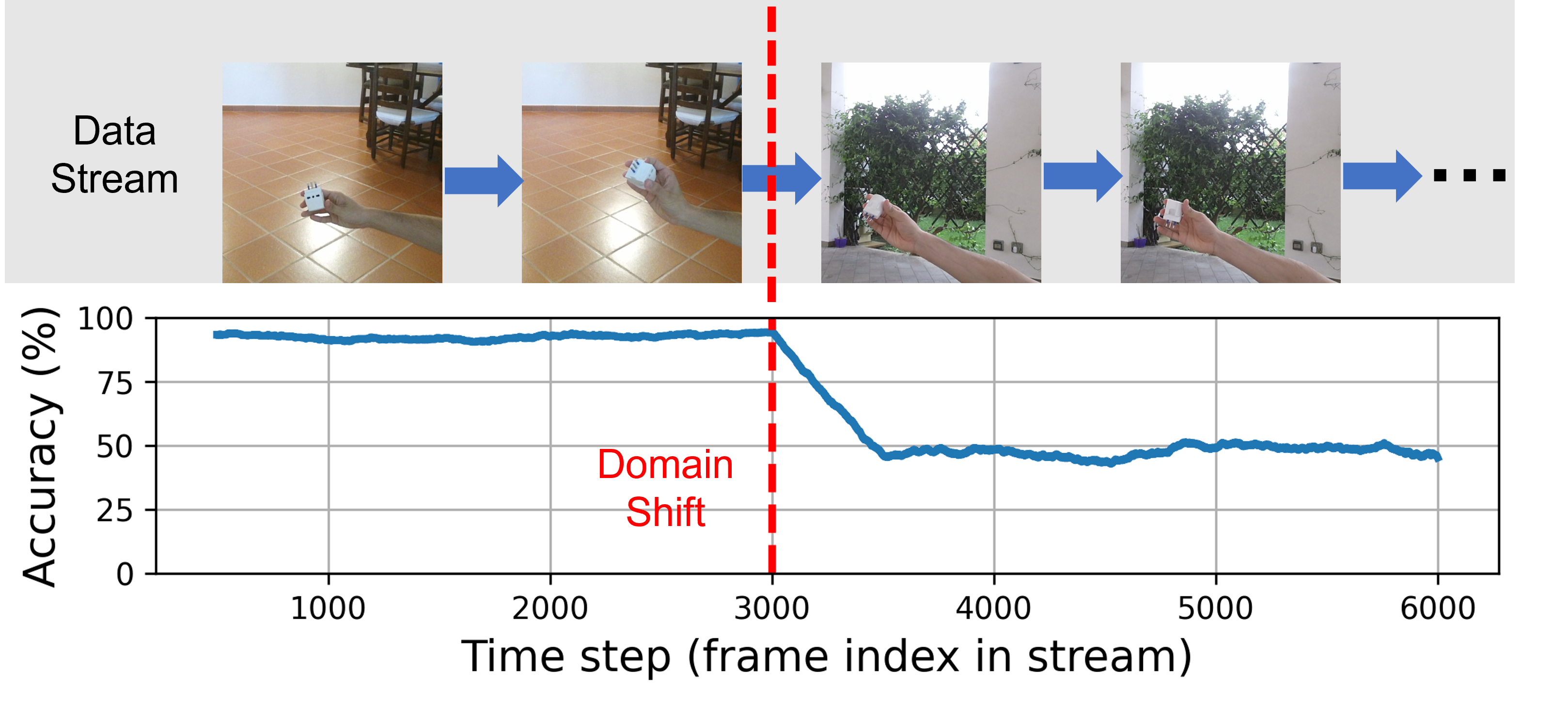}
  \vspace{-0.1in}
  \caption{Impact of domain shift: indoor $\rightarrow$ outdoor.}
  \label{fig: domain shift impact}
\vspace{-3mm}
\end{figure}

To illustrate the impact of domain shift, we conducted an experiment using CORe50~\cite{pmlr-v78-lomonaco17a}, a realistic continuous object recognition dataset for robots that contains indoor and outdoor sessions. As shown in \Cref{fig: domain shift impact}, we feed a continuous stream that transitions from an indoor session to an unseen outdoor session while keeping the object identity unchanged, and plot the rolling classification accuracy over time. The accuracy drops sharply from 94\% to 49\% after moving outdoor, even though the object identity remains unchanged. Since it is impractical to pretrain a model on all possible operating conditions, domain shift remains a key obstacle to robust embodied intelligence.

To address domain shift, \textbf{test-time adaptation (TTA)}~\cite{liang2025comprehensive} is an emerging solution. It adapts a pre-trained model \emph{during deployment} using only unlabeled streaming inputs, allowing the perception model to adjust to previously unseen conditions on the fly. Representative TTA methods~\cite{wang2020tent,niu2022eata,niu2023towards,hong2023mecta} usually minimize the prediction entropy (an unsupervised loss function) on unlabeled target data, encouraging the model to make more confident predictions in the new domain. Such methods typically operate batch by batch (a chunk of streaming inputs) and alternate between two steps: (1) an \textit{adaptation} step that updates model parameters via backpropagation,
using the entropy-minimization objective, 
followed by (2) an \textit{inference} step that produces predictions with the updated model. Repeating this procedure continuously adapts the model to each incoming batch, which is commonly referred to as \textbf{continual TTA} (C-TTA).

However, embodied devices are typically constrained by tight memory and energy budgets, making direct deployment of C-TTA impractical for two reasons. First, C-TTA relies on backpropagation during inference, which incurs substantial memory overhead, and this cost becomes especially prohibitive as batch size increases. Second, in realistic streaming scenarios, domain shifts are often intermittent: long stable periods exhibit only minor distribution changes, so adapting every batch is unnecessary and offers marginal benefit. For example, in \Cref{fig: domain shift impact}, accuracy stays stable for a long duration and drops noticeably only after approximately 3000 frames with scene transition, suggesting that continuously adapting during the stable period is wasteful.

In this paper, we first introduce a practical and efficient paradigm, referred to as \textbf{on-demand TTA}, to address key deployment challenges in embodied systems. Unlike \textbf{continual TTA (C-TTA)}, which adapts the model on every incoming batch, on-demand TTA activates adaptation only when a significant domain shift is detected, namely when the shift leads to an application-defined and unacceptable performance drop. By avoiding unnecessary updates during stable periods, on-demand TTA substantially reduces memory and energy overhead. Moreover, when no significant shift is detected, the system runs in a lightweight inference-only mode, enabling fast streaming predictions and making it well-suited for latency-sensitive applications.

However, realizing on-demand TTA on embodied devices presents several challenges. \textbf{First}, on-demand TTA requires continuous monitoring of the data distribution for every incoming sample/batch for potential domain shift detection. However, \textit{efficiently} quantifying the domain shift (or performance drop) without labels is challenging and remains underexplored in TTA literature. \textbf{Second}, differing from C-TTA, where the distribution of consecutive batches usually remains similar, on-demand TTA inherently deals with more severe shifts after a domain shift is detected. Existing C-TTA works have yet to address how to \textit{effectively} adapt under more significant shifts. \textbf{Third}, a notable limitation of many C-TTA methods is their reliance on \textit{large batch sizes} (therefore substantial memory consumption) to capture the new domain distribution~\cite{wang2020tent, niu2022eata}, which is impractical on embodied devices with limited onboard resources.

To tackle these challenges, we propose \textbf{\sysname}, a resource-efficient on-demand \textbf{TTA} framework designed for resource-constrained embodied devices. We drew three key insights from our observations and experimental studies to guide the design of \sysname. \textbf{First}, we observed that the entropy of model prediction can be used for detecting domain shifts. Based on this, we devised \textit{a novel lightweight domain shift detection mechanism} using exponential moving average (EMA) entropy to detect the domain shift. \textbf{Second}, we observe that the choice of source domain substantially affects post-adaptation performance (e.g., adapting from \textit{frost} to \textit{snow} yields better results than adapting from \textit{fog} to \textit{snow}). Instead of always adapting from the immediately preceding domain as in C-TTA, we propose a \textit{similar-domain selection} pipeline that identifies and selects the most relevant source domain for adaptation, leading to higher accuracy. \textbf{Third}, inspired by the insight that updating the Batch Normalization (BN) statistics and BN parameters consumes different amounts of memory and shows different sensitivity to batch sizes, we designed \textit{a decoupled BN update scheme} that adapts the BN statistics and BN parameters \textit{asynchronously} with different batch sizes, enabling effective model adaptation within a constrained memory budget.

We deploy \sysname on a Jetson Orin Nano, a portable computing platform commonly used in embodied systems, and evaluate it on four visual datasets spanning both object recognition and semantic segmentation tasks: object recognition with CIFAR10-C~\cite{hendrycks2019benchmarking}, ImageNet-C~\cite{hendrycks2019benchmarking}, CORe50~\cite{pmlr-v78-lomonaco17a}, and semantic segmentation with SHIFT~\cite{sun2022shift}. The experimental results show that \sysname achieves higher TTA accuracy while reducing energy consumption by 47.2\%, with comparable memory usage. Our comprehensive evaluation further highlights the robustness of \sysname, reinforcing its practical utility for real-world deployment under resource-constrained conditions.

Our contributions are summarized as follows:

\begin{itemize}[leftmargin=*, topsep=-15pt]

\item We introduce the concept of on-demand TTA for resource-constrained embodied visual systems. Our approach triggers adaptation only when significant domain shifts occur, thereby reducing adaptation frequency by up to 90\% over C-TTA methods while achieving the highest overall TTA accuracy across nearly all evaluation settings.

\item We present \sysname, a novel on-demand TTA framework for embodied devices. It comprises three core innovations: a lightweight domain shift detector, a source domain selection module, and a decoupled BN updating strategy.

    \item We implement \sysname on an embodied device and evaluate its performance across multiple datasets from various perspectives. Our results indicate that \sysname achieves higher TTA accuracy while reducing energy consumption by 47.2\% and latency by 6.7$\times$. 
\end{itemize}


\section{Preliminary}

\begin{figure}[t]
  \subfigure[]{
		\includegraphics[width = 0.45\textwidth]{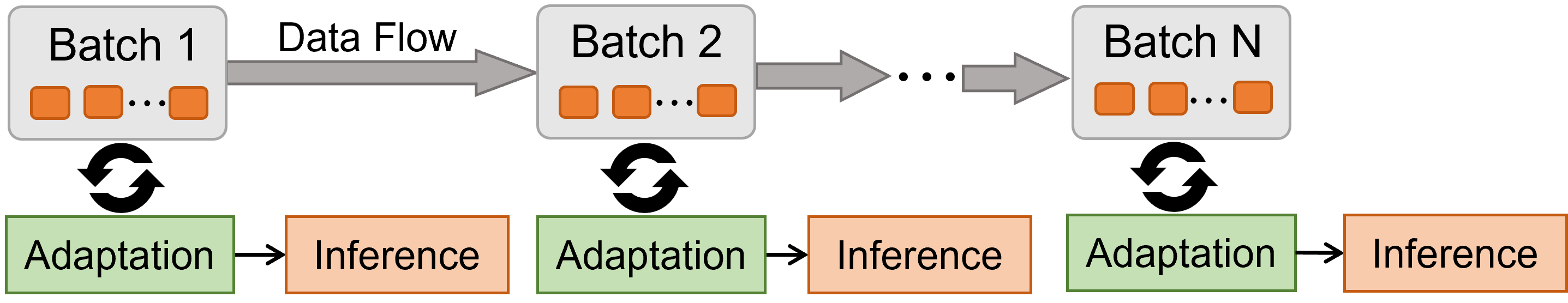}	
		\label{fig:continual-tta}} \subfigure[]{
		\includegraphics[width = 0.45\textwidth]{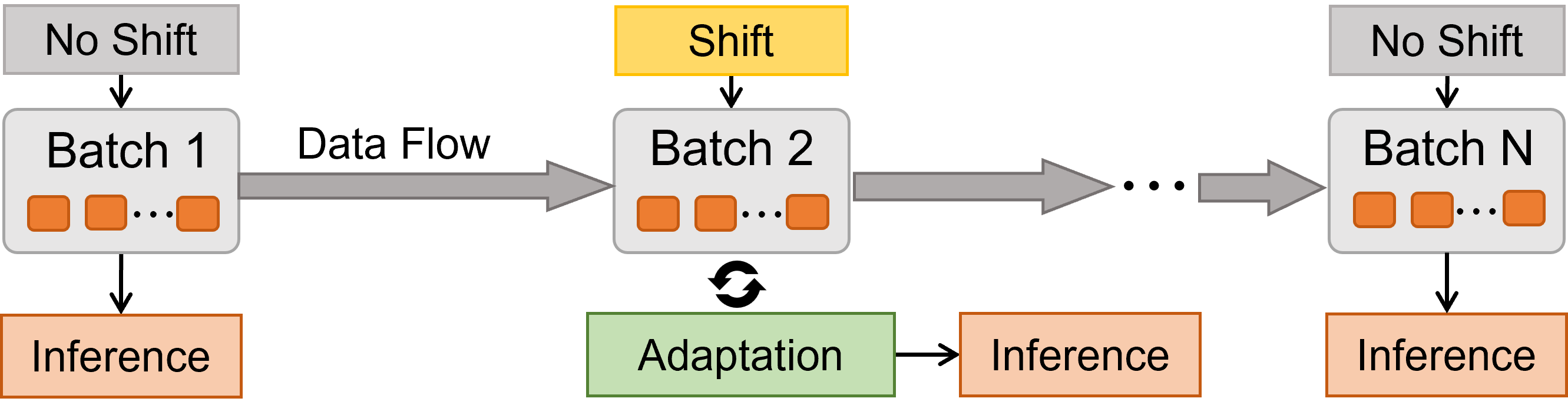}	
		\label{fig:on-demand-tta}}
  \vspace{-0.1in}
  \caption{Illustration of (a) continual and (b) on-demand test-time adaptation.}
  \label{fig: tta}
\vspace{-3mm}
\end{figure}

In this section, we present a detailed comparison between conventional continual TTA and the proposed on-demand TTA, followed by three key insights that inspire the design of \sysname. 




\subsection{Continual TTA vs. On-demand TTA}

\Cref{fig:continual-tta} depicts the process of existing TTA approaches, which perform two sequential steps for \textit{each incoming batch of new data}: an \textit{adaptation} step that updates certain layers (e.g., BN layers) by minimizing the entropy loss to align with the batch distribution, followed by an \textit{inference} step that produces predictions with the adapted model. This paradigm, referred to as \textbf{continual TTA} (C-TTA), although achieving optimal performance, is impractical for real-world deployment: adapting on every batch before inference incurs substantial latency and energy overhead even when the domain shift between consecutive batches is trivial.

Instead, a more practical TTA paradigm is to adapt the model only when a significant domain shift occurs and the current model suffers from an unacceptable performance degradation. As shown in \Cref{fig:on-demand-tta}, we introduce a new concept termed \textbf{on-demand TTA} for practical and efficient model adaptation during inference. On-demand TTA continuously monitors each incoming batch for potential domain shifts. If no shift is detected (e.g., Batch 1 and Batch N), the model proceeds with regular inference. However, once a domain shift is identified (e.g., Batch 2), the current model undergoes an adaptation process and performs inference on incoming data until the next domain shift is detected. Considering that domain shifts typically occur less frequently in real-world scenarios~\cite{liu2023lote}, on-demand TTA has the potential to significantly improve the utilization efficiency of on-device resources (e.g., computation and energy) by reducing the adaptation frequency, without sacrificing much accuracy.

\subsection{Challenges for On-demand TTA}
\label{sec: challenge}

Despite potential benefits, realizing on-demand TTA on embodied devices presents several additional challenges compared to C-TTA. \textbf{First (when to adapt)}, on-demand TTA triggers adaptation only when a substantial domain shift is detected and model performance drops. However, quantifying the performance drop without the labels is challenging. Additionally, since domain shift detection needs to be performed continuously on every input data, the detection process must be lightweight to conserve on-device resources. \textbf{Second (where to adapt from)}, unlike continual TTA, where the distribution captured by the current model and the distribution of the incoming data are often similar, on-demand TTA encounters a large distribution disparity when an adaptation is triggered. Therefore, it is unclear whether adapting from a largely deviated model will still be effective. \textbf{Third (how to adapt)}, existing C-TTA approaches rely on large batch sizes to accurately capture the distribution of new data, which consumes a significant amount of memory during adaptation (storing intermediate tensors during backpropagation). However, embodied devices generally have limited memory and need to process streaming data, making effective TTA with small batch sizes (even batch size = 1) challenging.

\begin{figure}[t]
  \subfigure[]{
		\includegraphics[width = 0.21\textwidth]{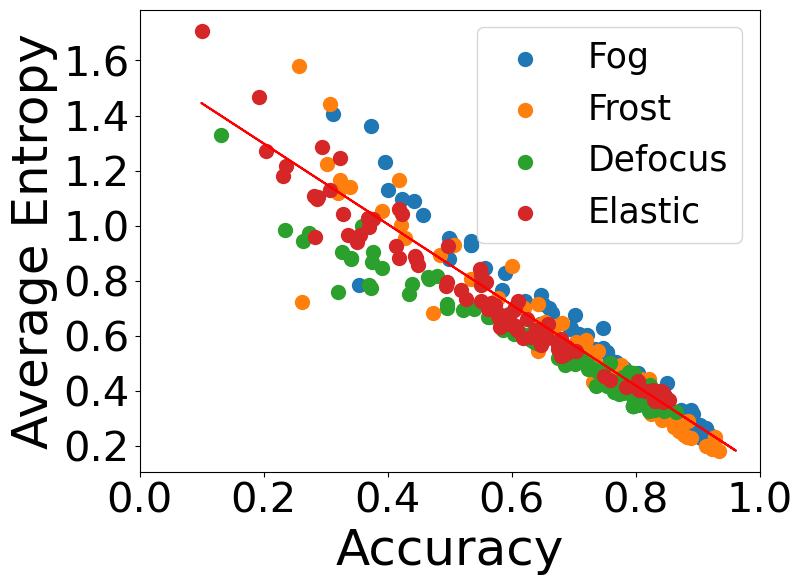}
		\label{fig: acc_entropy}} 
        \hspace{0.1in}\subfigure[]{
		\includegraphics[width = 0.21\textwidth]{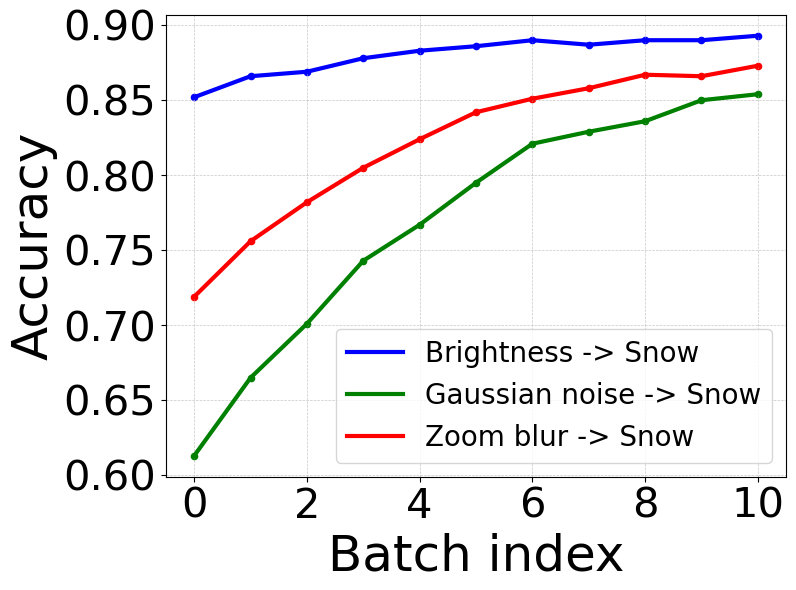}	
		\label{fig:similar_domain_bs4}}
  \vspace{-0.15in}
  \caption{(a) Correlation of accuracy and entropy, (b) adaptation to a target domain from different source domains.}
  \label{fig: fig1}
\vspace{-4mm}
\end{figure}
\subsection{Key Insights}
Before introducing our solutions, we first present some key insights that inspire our design. 


\subsubsection{Approximating accuracy with entropy}
\label{sec: insight1}
The first objective in on-demand TTA is to detect the occurrence of domain shifts in a \textit{lightweight} manner, as this needs to be performed continuously on all incoming data. Intuitively, one might consider tracking the accuracy drop during inference to identify such shifts. However, since the ground truth labels of the inference data are unavailable in practice, this method becomes infeasible. Inspired by entropy minimization~\cite{wang2020tent} in unsupervised learning, which improves model performance by reducing the entropy of predictions, we hypothesize that \textit{model performance is inversely correlated with entropy}. This insight leads us to explore using entropy as a potential metric to assess changes in model accuracy due to domain shifts.

To verify our hypothesis, we conducted an experiment using a pre-trained ResNet-50 model on CIFAR10-C, a common domain shift dataset comprising 15 different domains. We first adapted the source model to four selected domains using supervised learning to ensure effective adaptation. Then, we tested the adapted models on 75 subsets sampled from other domains, and we calculated the average accuracy and entropy in each subset. The results are shown in \Cref{fig: acc_entropy}, where each color represents the adapted model for each domain, and each scatter point denotes the accuracy and entropy of one subset. It is evident that entropy and accuracy are inversely correlated, with the relationship appearing nearly linear. \textbf{This inverse correlation between prediction entropy and model accuracy motivates us to approximate model performance by tracking the entropy of each input sample, without the need for ground-truth labels.}

\subsubsection{Adaptation from closer domain results in higher effectiveness}
\label{sec: insight2}

C-TTA always adapts the model from the previous domain, which may not be effective in on-demand TTA due to substantial distribution shifts. Based on the key observation that different domains exhibit varying degrees of similarity (e.g., in weather conditions, foggy and frost seem closer, compared to foggy and sunny), we hypothesize that \textit{the source domain (i.e., the domain before adaptation) can significantly impact the adaptation performance}. To test this, we adapted the model to the Snow domain (target domain) from three different source domains: Brightness, Zoom blur, and Gaussian noise. The accuracies of directly performing inference of the three source-domain models (i.e., models trained with the data from each domain only) on the Snow domain data were 85.2\%, 71.9\%, and 61.3\%, respectively, which indicates that Brightness is the closest to Snow, followed by Zoom blur and Gaussian noise.   

\Cref{fig:similar_domain_bs4} displays the adaptation accuracies with different numbers of batches. We can observe that adapting from a closer domain (i.e., Brightness) yields higher accuracy after adaptation, suggesting that 
the source domain indeed affects the adaptation performance. Moreover, starting from a closer domain leads to faster convergence: adapting from Brightness to Snow reaches a stable accuracy within four batches, whereas starting from Zoom Blur requires more than ten batches. Overall, these results suggest that \textbf{using a more similar source domain as the starting point provides a better initialization for adaptation, improving both the final adapted accuracy and the convergence speed.}

\begin{figure}[t]
  \subfigure[]{
		\includegraphics[width = 0.21\textwidth]{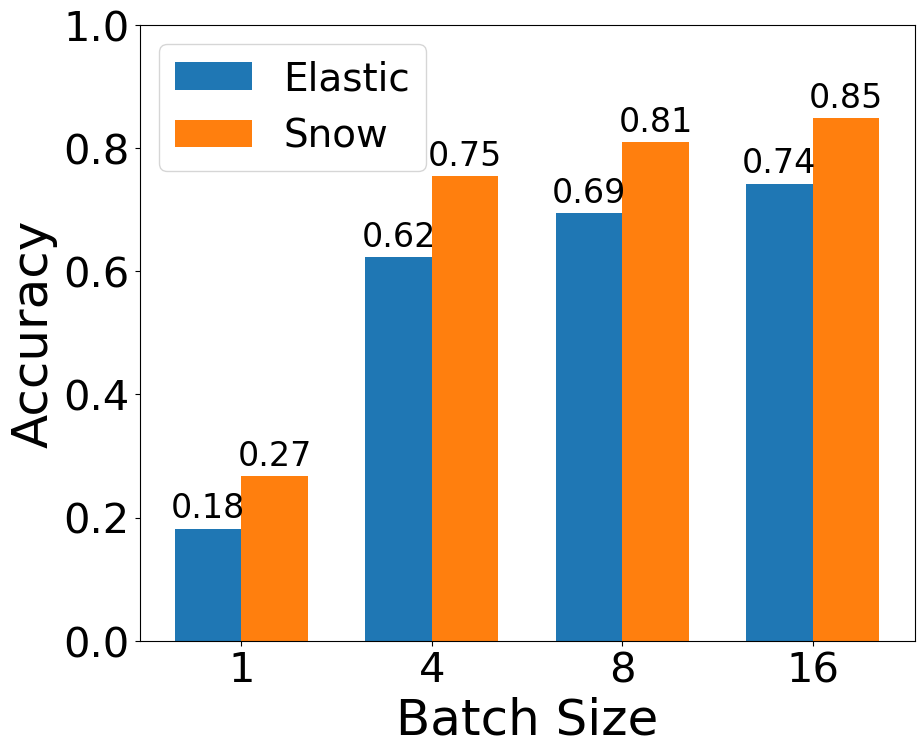}	
		\label{fig:bs_performance}} \hspace{0.1in}\subfigure[]{
		\includegraphics[width = 0.21\textwidth]{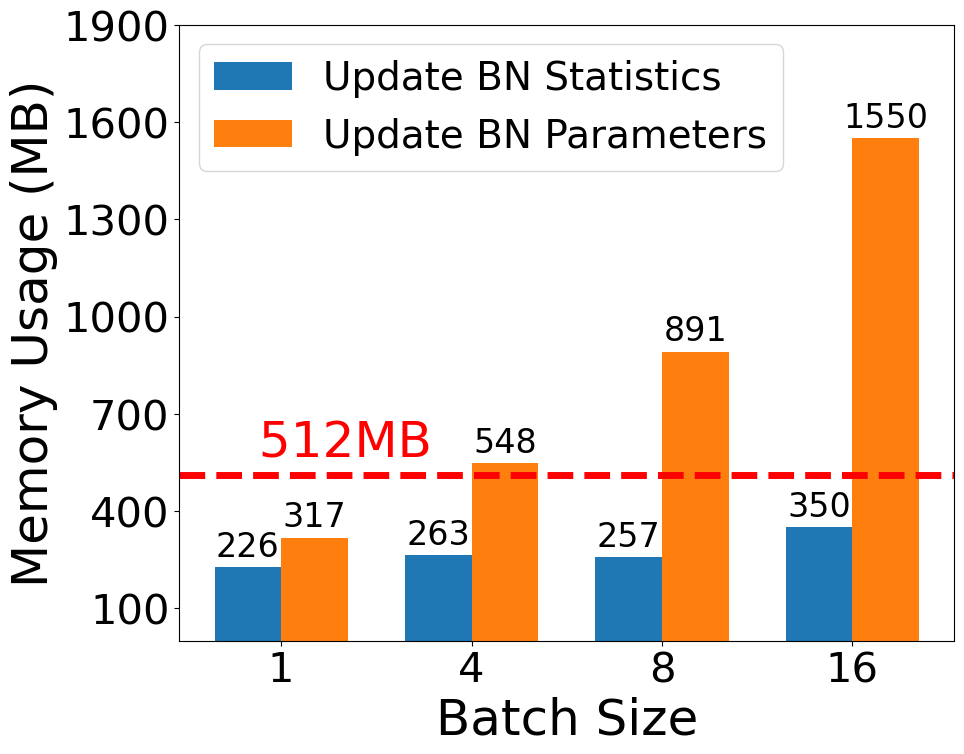}	
		\label{fig:memory_bn}}
  \vspace{-0.15in}
  \caption{(a) TTA performance under different batch sizes, (b) memory usage of updating BN statistics and parameters.}
  \label{fig/performance_memory}
  \vspace{-4mm}
\end{figure}

\subsubsection{Effectively updating BN statistics requires a large batch size yet incurs low memory usage}
\label{sec: insight3}

\textit{Most existing TTA approaches only adapt the BN layers rather than the entire model}, as it is sufficient to align the model with distribution shifts. However, these approaches rely on a large batch size (e.g., 64 in Tent~\cite{wang2020tent} and EATA~\cite{niu2022eata}) to capture the new domain's distribution. \Cref{fig:bs_performance} validates this by exemplifying the adaptation performance of EATA on the Elastic and Snow domain under different batch sizes, where smaller batch sizes yield poorer performance. On the other hand, larger batch sizes also demand significantly more memory to store intermediate activations and gradients for back-propagation, posing a challenge for memory-constrained embodied devices. For example, as shown in \Cref{fig:memory_bn}, with a batch size of 4, the memory requirement for updating BN parameters of ResNet-50 can easily exceed the available memory space of some small embodied devices such as Raspberry Pi Zero 2W with only 512MB DRAM.

To address this challenge, we delved into the detailed operation of BN layers during adaptation. Our detailed analysis reveals that \textit{BN statistics and BN affine parameters exhibit different sensitivities to batch size and impose distinct memory overheads}. We further analyzed the memory usage of updating BN statistics and BN parameters under different batch sizes in ResNet-50. As shown in \Cref{fig:memory_bn}, increasing the batch size results in significantly higher memory consumption when updating BN affine parameters, as this process requires backpropagation. In contrast, the memory usage for updating BN statistics remains relatively stable, since it only involves the forward pass. \textbf{This finding reveals that the memory overhead associated with large batch size primarily stems from updating BN parameters. Conversely, updating BN statistics, which is crucial for capturing data distributions, can be performed using large batch sizes without incurring substantial memory overhead.}

\section{\sysname Design}

\begin{figure}[t]
\centering
  \includegraphics[width = 0.45\textwidth]{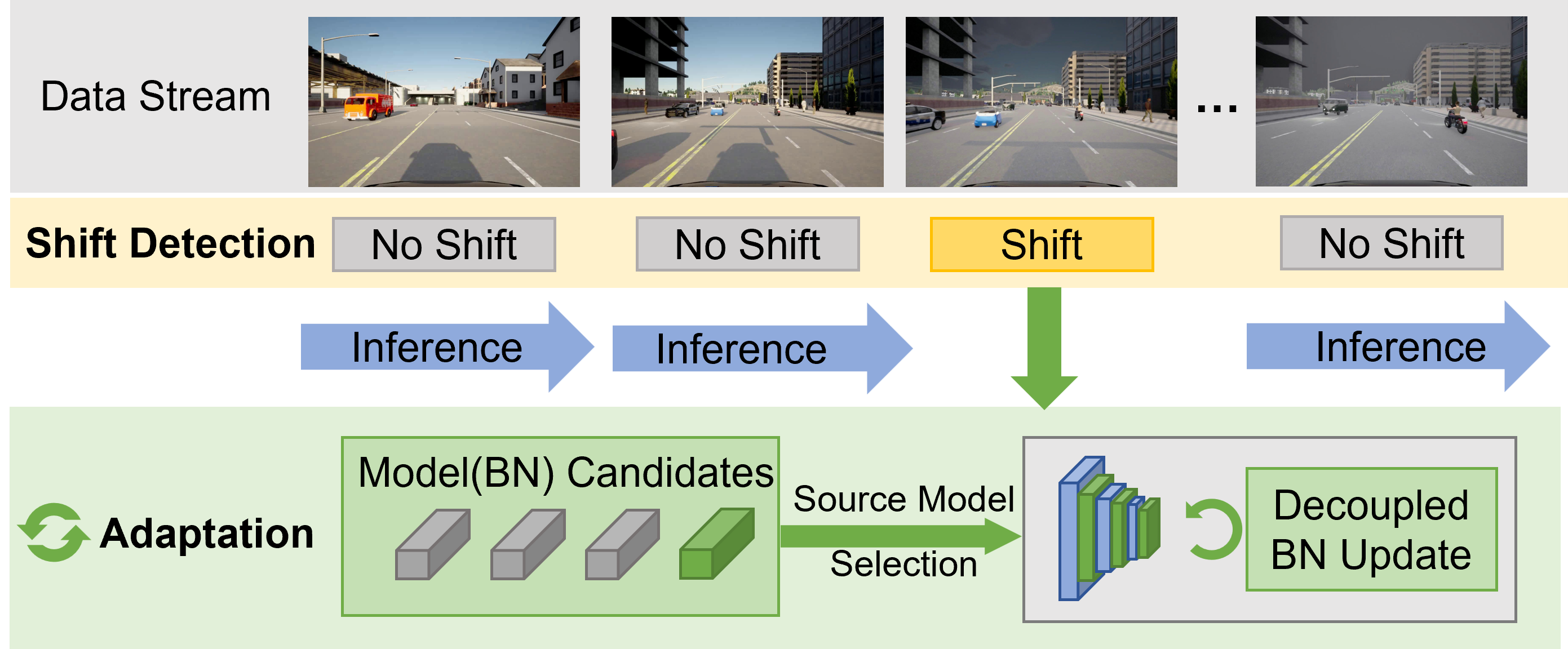}
  \vspace{-3mm}
  \caption{The overview of \sysname.}
  \label{fig: overview}
\vspace{-3mm}
\end{figure}

Based on the above insights, we propose \sysname, an on-demand TTA framework for resource-constrained embodied devices. As shown in \Cref{fig: overview}, \sysname comprises two modules: a lightweight \textit{domain shift detector} (Section~\ref{sec: shift detection}) that monitors incoming data and triggers adaptation only when needed, and a \textit{model adaptation} module that selects the closest source domain from a candidate pool (Section~\ref{sec: domain selection}) before applying a decoupled BN update (Section~\ref{sec: decoupled BN update}) to align the model with the new domain under memory constraints.

\subsection{Domain Shift Detection}
\label{sec: shift detection}

\begin{figure}[t]
  \subfigure[]{
		\includegraphics[width = 0.20\textwidth]{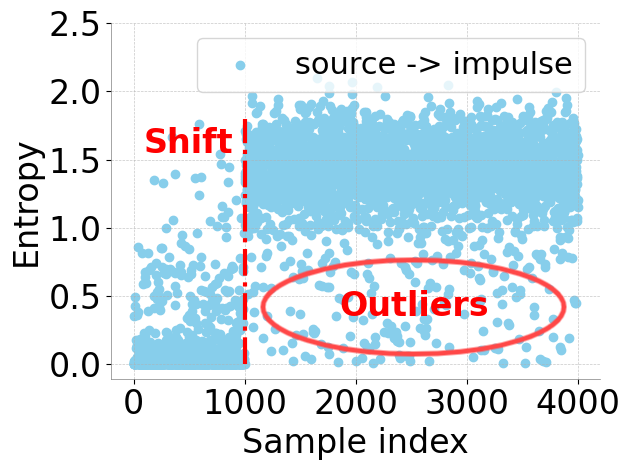}	
		\label{fig:entropy_curve}} \hspace{0.1in}\subfigure[]{
		\includegraphics[width = 0.20\textwidth]{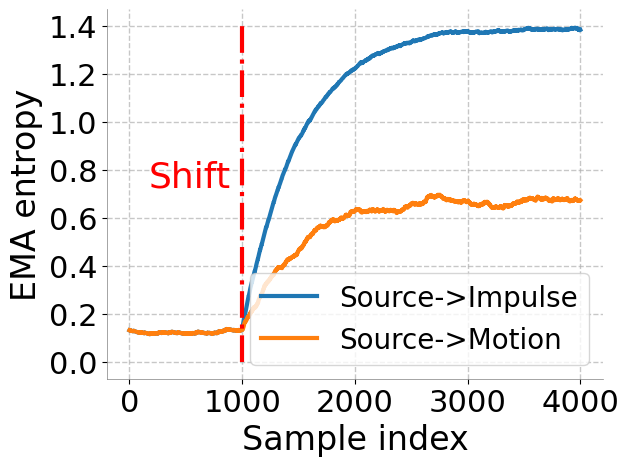}	
		\label{fig:ema_curve}}
  \vspace{-0.1in}
  \caption{Sample-wise domain shift detection using (a) entropy and (b) the proposed EMA entropy.}
  \label{fig/entropy_shift}
\vspace{-4mm}
\end{figure}

\subsubsection{EMA entropy calculation}


In \Cref{sec: insight1}, we presented the inverse correlation between model accuracy and average entropy. However, this correlation is demonstrated on a subset of samples (\textit{subset-wise}) and does not consistently hold in real-world scenarios where data is processed in a streaming manner. In such cases, DNNs operate on each individual sample, resulting in significant variations in \textit{sample-wise} entropy. Specifically, \Cref{fig:entropy_curve} illustrates the sample-wise entropy (denoted by each scatter) when the streaming samples shift from one domain (Source) to another domain (Impulse Noise).  
It is evident that while most samples in Impulse domain exhibit significantly higher entropy levels, some samples still present low entropy, hindering the direct detection of domain shifts using sample-wise entropy. 

To address this problem, we introduce an exponential moving average (EMA) strategy to smooth the sample-wise entropy by incorporating historical entropy and reducing the influence of a single sample. The formula for calculating the EMA entropy is as follows:
\textcolor{black}{
\begin{equation}
E_t = m \cdot E_{t-1} + (1 - m) \cdot x_t.
\label{eq:ema}
\end{equation}}
where \( E_t \) represents the EMA entropy at time \( t \), \( m \) denotes the momentum factor (with a value between 0 and 1), and \( x_t \) is the entropy value of the current input sample at time \( t \) that can be directly obtained from the inference process. The momentum \( m \) influences the stability of the EMA entropy and its sensitivity to domain shifts. A higher momentum value causes the current entropy to contribute minimally to the EMA entropy, resulting in a more stable curve but reducing sensitivity to domain shifts. In this paper, we empirically set the momentum to 0.995 to balance this trade-off. \Cref{fig:ema_curve} illustrates the EMA entropy curves for two adaptations, which clearly indicate the effectiveness of EMA entropy in detecting domain shifts.

\subsubsection{Shift determination}
\label{sec: shift determination}

After each adaptation, \sysname records the EMA entropy over the next few samples (e.g., 100) as the entropy baseline ($EMA_{base}$), which reflects the current model's capability on the adapted domain data. With the incoming data stream, \sysname calculates the sample-wise EMA entropy directly from the model inference outputs, referred to as $EMA_{sample}$. \textit{Notably, the extra computation from entropy to EMA entropy involves only simple multiplication and addition and therefore the proposed method is very lightweight}. Once $EMA_{sample}-EMA_{base}$ exceeds a threshold ($EMA_{thr}$), an adaptation is triggered. The threshold ($EMA_{thr}$) controls the trade-off between adaptation frequency and energy efficiency: a higher threshold tolerates more drift before triggering adaptation (saving energy), while a lower threshold adapts more readily for better accuracy at the cost of higher energy consumption. We provide more discussion of the detection sensitivity in Section~\ref{sec: ablation_detection}.

\textcolor{black}{To avoid manually selecting an absolute threshold for each dataset, we determine the threshold based on the historical distribution of EMA entropy under a stable data distribution. Specifically, we first compute the EMA entropy over a set of stable samples using the same momentum setting as during deployment. Let $\sigma_E$ denote the standard deviation of the resulting EMA entropy sequence. We then define the detection threshold as
\begin{equation}
    EMA_{\mathrm{thr}} = k \sigma_E,
\end{equation}
where $k$ is a dataset-independent scaling factor. Accordingly, a domain shift is detected when
\begin{equation}
    EMA_{\mathrm{sample}} - EMA_{\mathrm{base}} > k\sigma_E.
\end{equation}
This formulation scales the detection threshold according to the natural fluctuation of EMA entropy under a stable distribution, avoiding the need to manually specify different absolute thresholds for different datasets. In our experiments, we use a unified setting of $k=5$ across all datasets. The historical samples are used only to estimate $\sigma_E$, and no ground-truth labels are required for threshold selection.}

\subsection{Source Domain Selection}
\label{sec: domain selection}
As noted in \Cref{sec: insight2}, adapting from a closer domain can enhance both the accuracy and the convergence speed of adaptation. This observation motivates us to select the domain most similar to the new domain from a candidate pool before adaptation, rather than directly adapting from the last domain, referred to as source domain selection\footnote{Here the source domain refers to the domain before each adaptation.}. This process necessitates two essential steps: 1) constructing a pool with enough candidates to ensure effectiveness from the start (Section~\ref{sec:construction}) and during the runtime (Section~\ref{sec:progressive}), and 2) measuring the domain similarity to identify the candidate most closely aligned with the new domain (Section~\ref{sec: similar candidate selection}).

\subsubsection{Candidate pool construction}
\label{sec:construction}

\begin{figure}[t]
  \subfigure[]{
		\includegraphics[width = 0.21\textwidth]{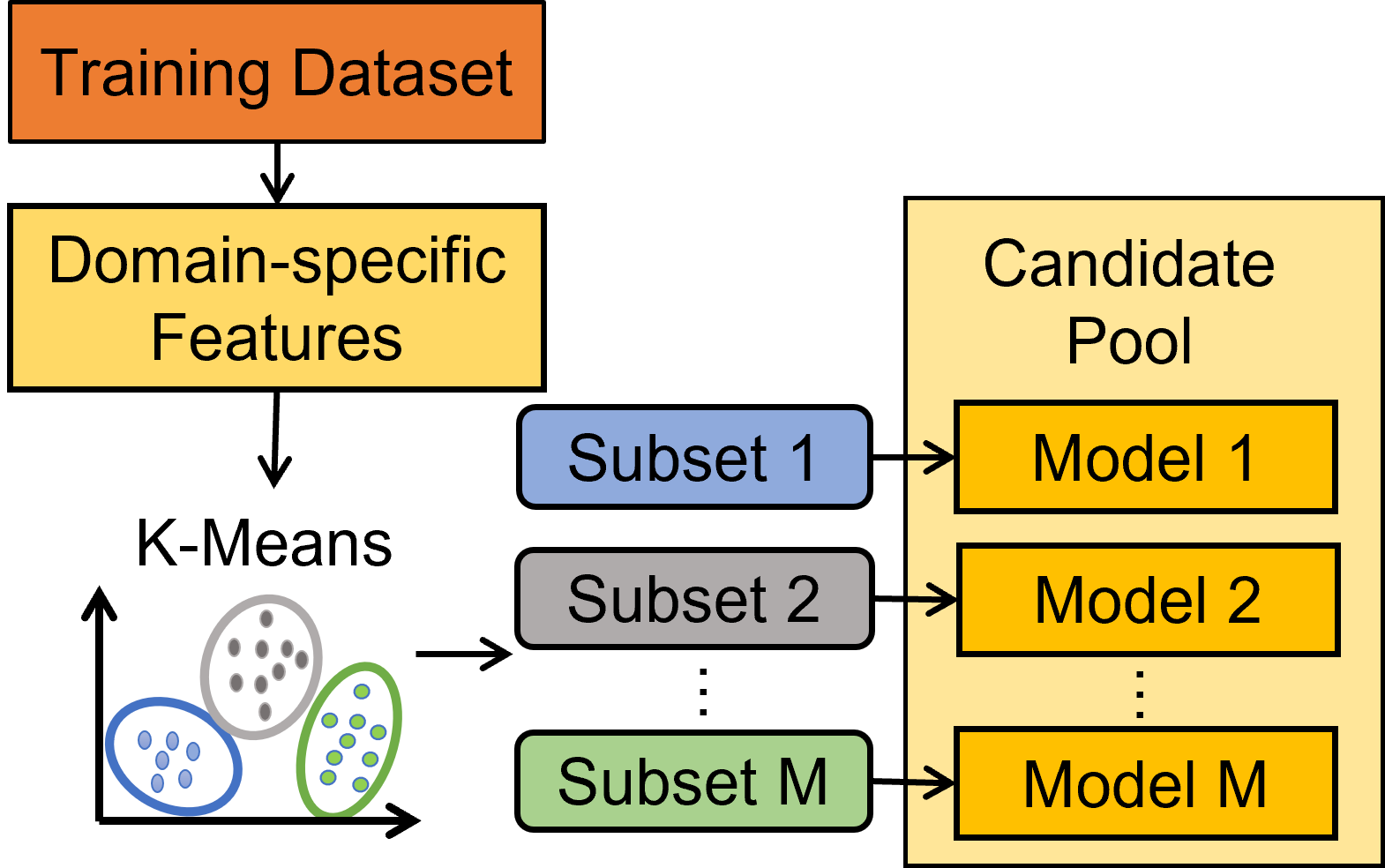}	
		\label{fig:fake}}
  \hspace{0.1in}\subfigure[]{
		\includegraphics[width = 0.21\textwidth]{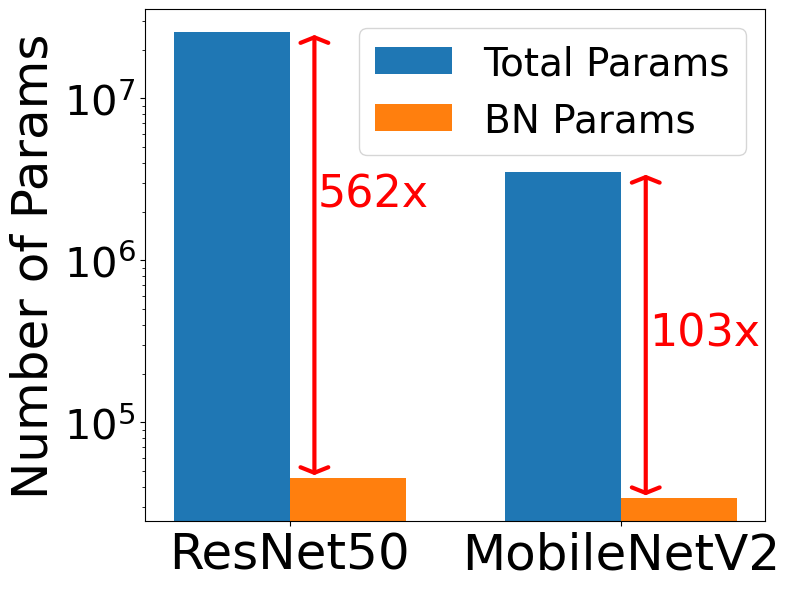}	
		\label{fig:storage}}

  \vspace{-0.1in}
  \caption{(a) Candidate domain construction, (b) storage comparison of saving only BN layers and the full model.}
  \label{fig/construction}
\vspace{-4mm}
\end{figure}

Considering that a domain essentially reflects the distribution of a specific set of data, the process of creating a domain candidate pool can be converted as generating multiple sets of data with diverse distributions. Additionally, training datasets typically contain dispersed distributions due to the varied data sources and collection conditions. Therefore, to construct the candidate pool, we propose to split the training dataset into multiple subsets and adapt the pre-trained model on each subset.


\Cref{fig:fake} depicts the detailed process\footnote{The candidate pool is constructed during model training, prior to online adaptation. This process does not require access to the training dataset during TTA.}. First, we extract features that can represent the domain characteristics (i.e., domain features) for each training sample. Specifically, given that BN statistics mainly capture data distribution, we feed each sample through the pre-trained model and utilize the output BN mean from the second BN layer as the domain features. We select the BN mean instead of the BN variance because the mean provides a stronger indication of domain shifts, while the variance primarily reflects distribution stability~\cite{niloy2024effective}. Shallow BN layers are preferred because they capture more domain-specific features, while deeper layers tend to focus on task-specific knowledge~\cite{niloy2024effective}. Meanwhile, we avoid using the first BN layer as it processes very primitive features immediately after the initial convolution, which are less informative for domain characterization. 
Second, based on these domain features, we cluster the training samples into $m$ subsets using the K-Means algorithm~\cite{macqueen1967some}, with each cluster representing a domain. Third, we adapt the pre-trained model on each subset by freezing all the layers except for the BN layers (same as conventional TTA), resulting in $m$ domain candidates in the pool by saving the BN layers. Unlike TTA, this adaptation is supervised using known training labels to ensure accurate BN updates.

Alternatively, when source data is unavailable, \sysname can progressively build the candidate pool at runtime\label{sec:progressive}: starting from the source model as the initial candidate, each newly adapted model is stored (saving only its BN parameters) as an additional candidate for future domains. We evaluate this progressive construction scheme in Section~\ref{sec:ablation_selection}, and the two strategies can be combined to build a stronger candidate pool.

\textit{Overhead of storing domain candidates:} As TTA only adapts BN layers, storing the candidate models requires only storing the BN layers. To understand the storage overhead, we measured the storage requirements for saving just the BN layers compared to the entire model. \Cref{fig:storage} demonstrates that BN parameters account for only 1/562 and 1/103 of the total parameters (translating to 480 KB and 210 KB) in ResNet-50 and MobileNetV2, respectively. This finding suggests that the overhead of storing multiple domain candidates (e.g., 100) is negligible in terms of memory consumption, promoting the feasibility of the proposed domain selection strategy.  

\subsubsection{Similar candidate selection}
\label{sec: similar candidate selection}

After constructing a candidate pool, the next step is to select the candidate domain most similar to the incoming target domain. This similarity estimation relies on accurately representing the new domain using BN statistics extracted via the source model. However, the source model typically performs poorly on shifted domains due to BN mismatches, which degrades the quality of extracted features (see \Cref{sec: insight3}).

Specifically, we cache \( N \) samples from the new domain (e.g., \( N = 128 \)) and process them in small batches (e.g., batch size 16) through the source model to extract domain features. Before extraction, we update the BN statistics via a forward pass to align the model with the batch distribution\footnote{The process only requires updating BN statistics via a forward pass without backpropagation, resulting in low peak memory consumption, as illustrated in \Cref{fig:memory_bn}.}. Then, for each batch \( \mathcal{B}_k \), we compute the BN mean from the second BN layer and average over the \( K \) cached batches to obtain the final domain representation:

\vspace{-1mm}
\begin{equation}
\mu_{\text{domain}} = \frac{1}{K} \sum_{k=1}^K \left( \frac{1}{|\mathcal{B}_k|} \sum_{i \in \mathcal{B}_k} \phi(x_i) \right),
\label{domain_representation}
\end{equation}

where \( \phi(x_i) \) denotes the intermediate output of sample \( i \) in batch \( \mathcal{B}_k \) after the second BN layer, which we choose because it exhibits the highest correlation with domain similarity, as evaluated in Section~\ref{sec:ablation_selection}.

To identify the most similar candidate model from the pool, we compute the L2 distance between \( \mu_{\text{domain}} \) and each candidate’s stored BN mean \( \mu_c \), and select the closest one:

\begin{equation}
c^* = \arg\min_c \|\mu_{\text{domain}} - \mu_c\|_2.
\label{eq:candidate_selection}
\end{equation}

Here, \( c^* \) denotes the candidate whose BN statistics are most similar to the current domain representation. This selected candidate serves as the initialization point for the subsequent adaptation process. We further evaluate the effectiveness of this selection strategy in Section~\ref{sec:ablation_selection}.

\subsection{Decoupled BN Update}
\label{sec: decoupled BN update}

While similar candidate selection improves the performance of on-demand TTA, the challenge related to small batch sizes on resource-constrained embodied devices still persists. Recall that each BN layer contains two types of quantities: \emph{BN statistics} (running mean/variance for normalization) and \emph{BN parameters} (learnable affine scale $\gamma$ and shift $\beta$); updating statistics requires only a forward pass and is sensitive to batch size, whereas updating parameters via backpropagation is more robust to small batches~\cite{wang2020tent,niu2022eata,niu2023towards,hong2023mecta}. Inspired by the findings in \Cref{sec: insight3}, we propose a \textit{decoupled BN update} strategy: BN statistics are first updated with a larger batch size during a forward pass, followed by BN parameter fine-tuning with a small batch size during backpropagation.

\subsubsection{BN statistics update}
\label{sec: BN stats update}

Having cached $N$ samples for similar model selection, we efficiently reuse these samples to adapt the BN layers. Specifically, the samples are also split into batches of size 16. But unlike similar domain selection, which updates BN statistics on the previous poor model by averaging over batches, here the BN statistics are updated on the selected candidate model in a batch-by-batch manner.  
In detail, we employ an Exponential Moving Average (EMA) approach to integrate the BN statistics of the current batch with historical BN statistics, defined as follows:
\begin{equation}
S_k = (1 - m) \cdot S_{k-1} + m \cdot B_k \label{eq: running_mean}
\end{equation}

where \( S_k \) represents the integrated BN statistics at batch \( k \), \( m \) is the momentum factor, \( S_{k-1} \) is the BN statistics integrated from the previous batch, and \( B_k \) represents the BN statistics of the current batch. Particularly, \( S_0 \) is the BN statistics of the selected candidate model. 
The proposed EMA-based updating scheme allows us to seamlessly integrate the similar source model with the new domain samples, ensuring an accurate and robust alignment of both source and new domain distributions.

\subsubsection{BN parameters update}

After updating the BN statistics, the selected model already captures the distribution of the new data, but the BN parameters still need to be fine-tuned accordingly through back-propagation. To fit into the limited on-device memory (e.g., 512MB as plotted in~\Cref{fig:memory_bn}), we aim to update the BN parameters with a small batch size (e.g., 1). However, back-propagation using a single sample is challenging due to the inherent instability of unsupervised learning~\cite{niu2023towards}. To achieve stable fine-tuning, following EATA-C~\cite{tan2025uncertainty} and SAR~\cite{niu2023towards}, we filter out high-entropy samples when calculating the entropy loss, since such low-confidence predictions introduce noisy gradients that destabilize the unsupervised adaptation.

Therefore, the overall loss function is defined as:
\begin{equation}
\mathcal{L}_{\mathrm{ent}} \;=\;
\frac{1}{|\mathcal{S}|}\sum_{i\in \mathcal{S}} 
\Big( - \sum_{c=1}^{C} p_{i,c}\log p_{i,c} \Big),
\mathcal{S}=\{\, i \mid H(p_i) < \tau \,\},
\label{eq:filtered_entropy}
\end{equation}

In \Cref{eq:filtered_entropy}, $(-\sum_{c=1}^{C} p_{i,c}\log p_{i,c})$ is the prediction entropy. We compute the entropy-minimization loss only over the reliable subset $\mathcal{S}$, defined as samples whose entropy falls below a threshold $\tau$. Following EATA-C~\cite{tan2025uncertainty} and SAR~\cite{niu2023towards}, we set $\tau = 0.4\log(C)$, where $C$ is the number of task classes. \textcolor{black}{For the cached samples used in adaptation, the predictions produced during the forward pass for computing the BN-parameter adaptation loss are directly recorded as their final outputs. Therefore, no additional inference pass is performed after the parameter update.}

\section{Evaluation Setup}
\label{sec: implementation}

\noindent\textbf{Device.} We implement \sysname on Jetson Orin Nano (8GB RAM) running PyTorch 2.0 on Ubuntu 20.04. \textcolor{black}{Classification at batch sizes 1 and 16 is evaluated on Jetson Orin Nano whenever the method fits within the device memory; otherwise, the corresponding result is measured on the server. Batch size 64 is evaluated on the server for all methods.}

\noindent\textbf{Datasets.} We evaluate on four datasets. \textbf{CIFAR10-C} and \textbf{ImageNet-C}~\cite{hendrycks2019benchmarking} are corrupted variants of CIFAR10~\cite{krizhevsky2009learning} and ImageNet~\cite{deng2009imagenet} with 15 corruption types at severity level 5. \textbf{CORe50}~\cite{pmlr-v78-lomonaco17a} is a streaming object recognition dataset collected across 11 sessions (8 indoor, 3 outdoor); we train on indoor sessions and run TTA on the three outdoor sessions as unseen target domains. \textbf{SHIFT}~\cite{sun2022shift} is an autonomous driving segmentation dataset with day$\rightarrow$night, clear$\rightarrow$foggy, and clear$\rightarrow$rainy shifts; we follow~\cite{sun2022shift} and resize images to 640$\times$400 to reduce on-device overhead.

\noindent\textbf{Baselines.} We compare against representative TTA methods: \textbf{SAR}~\cite{niu2023towards} replaces BN with group normalization for batch-size robustness (using a TIMM ResNet-50-GN~\cite{rw2019timm}); \textbf{MECTA}~\cite{hong2023mecta} reduces memory via selective parameter updates through a redesigned normalization layer; \textbf{ROID}~\cite{marsden2024roid} uses certainty/diversity-based weighting with continual weight averaging; \textbf{L-TTA}~\cite{shin2024tta} adapts only lightweight stem layers to minimize updated tensors; \textbf{SURGEON}~\cite{ma2025surgeon} dynamically prunes activations for a controllable accuracy--memory trade-off; and \textbf{EATA-C}~\cite{tan2025uncertainty} extends EATA~\cite{niu2022eata} with uncertainty-aware sample filtering and calibration; \textcolor{black}{RoTTA~\cite{yuan2023robust} maintains a category-balanced memory bank to improve continual adaptation under dynamic test streams. We include RoTTA as a representative memory-bank-based TTA method to provide a direct comparison with cache-aware adaptation.}

\noindent\textbf{Adaptation Details.}\label{subsec:adaptation_details} For object recognition we use ResNet-50~\cite{he2016deep}; for segmentation, DeepLabV3~\cite{chen2017rethinking} with a ResNet-50 backbone (metric: mIoU). Baselines use their default settings. \textcolor{black}{Unless otherwise specified, the reported batch size refers to both the inference batch size and the batch size used for gradient-based adaptation. EmbodiTTA uses a larger batch for the decoupled BN-statistics update; however, this step is forward-only and consumes substantially less memory than backpropagation. As shown in Fig.~\ref{fig:memory_bn}, its memory usage remains below that of the gradient-based update even with a larger batch size. Hence, under the same peak-memory budget, the backpropagation batch size remains the dominant memory constraint and provides a fair basis for comparison. Cached samples are used for domain selection and BN-statistics estimation rather than increasing the backpropagation batch.} For \sysname, we use 128/512/512 adaptation samples on CIFAR10-C/ImageNet-C/CORe50 at batch size 1, and 512 samples for batch sizes 16 and 64. Learning rates are $1\times10^{-4}$ (CIFAR10-C, SHIFT) and $1\times10^{-5}$ (ImageNet-C). For shift detection, we determine the threshold as
$\mathrm{EMA}_{\mathrm{thr}}=k\sigma_E$, where $\sigma_E$ is estimated from the EMA entropy under a stable historical distribution. We use a unified setting of $k=5$ across all datasets without using target labels.

\section{Evaluation Results}

%
\subsection{Adaptation Performance}\label{subsec:main results}

\begin{table*}[]
\caption{Comparison of accuracy (\%) on CIFAR10-C and ImageNet-C using ResNet-50. Memory is measured on Jetson Orin Nano whenever the method fits on the device; otherwise, it is measured on a server. Batch size 64 is measured on a server for all methods due to Jetson memory limitations.}
\centering
\label{tab:overall_results}
\resizebox{0.85\textwidth}{!}{%
\begin{tabular}{@{}c|ccccccccc@{}}
\toprule
 & \multicolumn{3}{c}{Batch Size = 1} & \multicolumn{3}{c}{Batch Size = 16} & \multicolumn{3}{c}{Batch Size = 64} \\ \cmidrule(l){2-10} 
 & \multicolumn{2}{c|}{Avg. Acc.(\%)} & \multicolumn{1}{c|}{} & \multicolumn{2}{c|}{Avg. Acc.(\%)} & \multicolumn{1}{c|}{} & \multicolumn{2}{c|}{Avg. Acc.(\%)} &  \\ \cmidrule(lr){2-3} \cmidrule(lr){5-6} \cmidrule(lr){8-9}
\multirow{-3}{*}{Method} & CIFAR10-C & \multicolumn{1}{c|}{ImageNet-C} & \multicolumn{1}{c|}{\multirow{-2}{*}{\begin{tabular}[c]{@{}c@{}}Memory\\ (MB)\end{tabular}}} & CIFAR10-C & \multicolumn{1}{c|}{ImageNet-C} & \multicolumn{1}{c|}{\multirow{-2}{*}{\begin{tabular}[c]{@{}c@{}}Memory\\ (MB)\end{tabular}}} & CIFAR10-C & \multicolumn{1}{c|}{ImageNet-C} & \multirow{-2}{*}{\begin{tabular}[c]{@{}c@{}}Memory\\ (MB)\end{tabular}} \\ \midrule
Source & 59.5 & \multicolumn{1}{c|}{26.9} & \multicolumn{1}{c|}{242} & 59.5 & \multicolumn{1}{c|}{26.9} & \multicolumn{1}{c|}{358} & 59.5 & \multicolumn{1}{c|}{26.9} & 809 \\
SAR & 68.3 & \multicolumn{1}{c|}{\textbf{35.1}} & \multicolumn{1}{c|}{429} & 70.4 & \multicolumn{1}{c|}{35.2} & \multicolumn{1}{c|}{1723} & 69.2 & \multicolumn{1}{c|}{37.3} & 5780 \\
MECTA & 65.0 & \multicolumn{1}{c|}{12.0} & \multicolumn{1}{c|}{378} & 81.3 & \multicolumn{1}{c|}{33.2} & \multicolumn{1}{c|}{1231} & 81.3 & \multicolumn{1}{c|}{33.7} & 3836 \\
ROID & 10.1 & \multicolumn{1}{c|}{0.4} & \multicolumn{1}{c|}{698} & 79.5 & \multicolumn{1}{c|}{32.5} & \multicolumn{1}{c|}{2850} & 82.8 & \multicolumn{1}{c|}{37.8} & 9840 \\
L-TTA & 10.1 & \multicolumn{1}{c|}{0.1} & \multicolumn{1}{c|}{449} & 76.8 & \multicolumn{1}{c|}{26.6} & \multicolumn{1}{c|}{1250} & 81.2 & \multicolumn{1}{c|}{34.5} & 3580 \\
SURGEON & 10.0 & \multicolumn{1}{c|}{0.4} & \multicolumn{1}{c|}{454} & 74.5 & \multicolumn{1}{c|}{9.97} & \multicolumn{1}{c|}{1750} & 82.0 & \multicolumn{1}{c|}{36.3} & 5882 \\
EATA-C & 23.7 & \multicolumn{1}{c|}{1.0} & \multicolumn{1}{c|}{506} & 77.4 & \multicolumn{1}{c|}{31.9} & \multicolumn{1}{c|}{1728} & 81.3 & \multicolumn{1}{c|}{38.6} & 6228 \\
\textcolor{black}{RoTTA} 
& \textcolor{black}{66.2} 
& \multicolumn{1}{c|}{\textcolor{black}{32.8}} 
& \multicolumn{1}{c|}{\textcolor{black}{2829}} 
& \textcolor{black}{82.9} 
& \multicolumn{1}{c|}{\textcolor{black}{33.9}} 
& \multicolumn{1}{c|}{\textcolor{black}{11328}} 
& \textcolor{black}{81.9} 
& \multicolumn{1}{c|}{\textcolor{black}{34.9}} 
& \textcolor{black}{22588} \\
\rowcolor[HTML]{FFCCC9} 
Ours & \textbf{78.0} & \multicolumn{1}{c|}{\cellcolor[HTML]{FFCCC9}33.8} & \multicolumn{1}{c|}{\cellcolor[HTML]{FFCCC9}414} & \textbf{83.0} & \multicolumn{1}{c|}{\cellcolor[HTML]{FFCCC9}\textbf{37.4}} & \multicolumn{1}{c|}{\cellcolor[HTML]{FFCCC9}1632} & \textbf{84.9} & \multicolumn{1}{c|}{\cellcolor[HTML]{FFCCC9}\textbf{40.4}} & 5653 \\ \bottomrule
\end{tabular}%
}
\vspace{-5mm}
\end{table*}

\subsubsection{CIFAR10-C and ImageNet-C} \Cref{tab:overall_results} shows that \sysname achieves significant accuracy gains across all batch sizes. Notably, it achieves strong batch-size-1 performance among BN-based methods. It is critical for memory-constrained embodied devices—while SAR (GN-based) underperforms at larger batches. Memory-wise, \sysname achieves peak GPU usage comparable to EATA-C and SAR. Although MECTA and L-TTA can reduce memory at the same batch size through sparse or optimized backpropagation, their accuracy degrades substantially at batch size 1. To reach accuracy comparable to \sysname at batch size 1 on ImageNet-C (33.8\%, 414 MB), MECTA must increase the batch size to 16, where it consumes 1231 MB. This highlights that \sysname achieves a better accuracy--memory trade-off under small-batch embodied deployment.\footnote{Memory refers to peak GPU usage across the entire process including model loading, backpropagation, and runtime operations.} \textcolor{black}{Compared with RoTTA, a representative memory-bank-based TTA method,
\sysname achieves a substantially better accuracy--memory trade-off.
RoTTA maintains historical target samples and uses the memory bank for adaptation, which leads to considerable memory overhead
as the adaptation batch size increases. In contrast, the cached samples
in \sysname are mainly used to obtain more reliable target-domain
statistics and update BN statistics through forward-only operations,
while the gradient-based update is still performed with a small
backpropagation batch. Consequently, \sysname can exploit information
from multiple observed samples without incurring the memory cost of
backpropagating through a large cached batch.}

\begin{table}[t]
\centering
\caption{Comparison of accuracy (\%) on CORe50 dataset using ResNet-50.}
\label{tab:core50results}
\resizebox{0.8\linewidth}{!}{%
\begin{tabular}{@{}c|ccc@{}}
\toprule
Method & Batch size = 1 & Batch size = 16 & Batch size = 64 \\ \midrule
Source & 69.9 & 69.9 & 69.9 \\
SAR & 76.8 & 75.9 & 74.6 \\
MECTA & 62.7 & 83.4 & 84.2 \\
ROID & 6.2 & 81.3 & 84.3 \\
L-TTA & 2.2 & 83.0 & 83.3 \\
SURGEON & 2.9 & 79.5 & 80.4 \\
EATA-C & 6.7 & 73.4 & 78.0 \\
\rowcolor[HTML]{FFCCC9}
\textbf{Ours} & \textbf{83.7} & \textbf{84.0} & \textbf{84.5} \\ \bottomrule
\end{tabular}%
}
\vspace{-3mm}
\end{table}

\begin{table}[t]
\centering
\caption{Comparison of Adaptation mIoU (\%) on SHIFT using DeepLabV3.}
\label{tab:shift}
\resizebox{0.8\linewidth}{!}{%
\begin{tabular}{@{}c|ccc|c@{}}
\toprule
 & \multicolumn{3}{c|}{SHIFT types} &  \\ \cmidrule(lr){2-4}
Method & Day$\rightarrow$Night & Clear$\rightarrow$Foggy & Clear$\rightarrow$Rainy & Avg. \\ \midrule
Source  & 27.3 & 17.7 & 11.0 & 18.7 \\
SAR     & 23.6 & 7.6  & 4.2  & 11.8 \\
MECTA   & 24.2 & 20.6 & 15.4 & 20.1 \\
SURGEON & 30.1 & 23.6 & 18.2 & 24.0 \\
EATA-C  & 24.7 & 21.5 & 16.7 & 20.9 \\
\rowcolor[HTML]{FFCCC9}
\textbf{Ours} & \textbf{31.1} & \textbf{25.2} & \textbf{19.1} & \textbf{25.1} \\ \bottomrule
\end{tabular}%
}
\vspace{-3mm}
\end{table}

\subsubsection{CORe50} \Cref{tab:core50results} shows that \sysname achieves the best performance across all batch-size settings, reaching 83.7\% at batch size 1 and substantially outperforming the strongest baseline, SAR (76.8\%). Many baselines collapse at batch size 1 (e.g., ROID: 6.2\%, L-TTA: 2.2\%), highlighting the gap between conventional TTA evaluation and embodied deployment where small batches are often unavoidable. \sysname remains stable across all batch sizes.

\subsubsection{SHIFT} \Cref{tab:shift} shows that \sysname consistently outperforms all compared baselines across all shift types (avg.\ mIoU 25.1\% vs.\ 24.0\% for the next best, SURGEON).\footnote{ROID and L-TTA are not included in the SHIFT evaluation as their normalization designs are tightly coupled to classification heads and do not support semantic segmentation.} Notably, SAR—which achieves competitive recognition accuracy—performs worst on segmentation (avg.\ mIoU 11.8\%), underscoring the task-sensitivity of normalization design choices.

\begin{table}[]
\caption{Latency for processing domain data sequences of varying lengths under batch size of 1 on CIFAR10-C.}
\label{tab:latency}
\resizebox{0.49\textwidth}{!}{%
\begin{tabular}{@{}c|cccccc@{}}
\toprule
Lengths & 1000 & 3000 & 5000 & 7000 & 9000 & 10000 \\ \midrule
SAR & 138 (1.2$\times$) & 414 & 690 & 965 & 1241 & 1379 (1.2$\times$) \\
MECTA & 160 (1.0$\times$) & 480 & 800 & 1120 & 1440 & 1600 (1.0$\times$) \\
ROID & 75 (2.1$\times$) & 226 & 376 & 526 & 676 & 749 (2.1$\times$) \\
L-TTA & 50 (3.2$\times$) & 143 & 240 & 333 & 424 & 477 (3.4$\times$) \\
SURGEON & 78 (2.1$\times$) & 227 & 376 & 531 & 678 & 770 (2.1$\times$) \\
EATA-C & 47 (3.4$\times$) & 136 & 227 & 317 & 408 & 453 (3.5$\times$) \\
Ours & 47 (3.4$\times$) & 90 & 132 & 175 & 218 & 239 (6.7$\times$) \\ \bottomrule
\end{tabular}%
}
\end{table}

\subsubsection{Latency and Energy} 
\begin{figure}[t]
\centering
  \subfigure[]{
		\includegraphics[width = 0.21\textwidth]{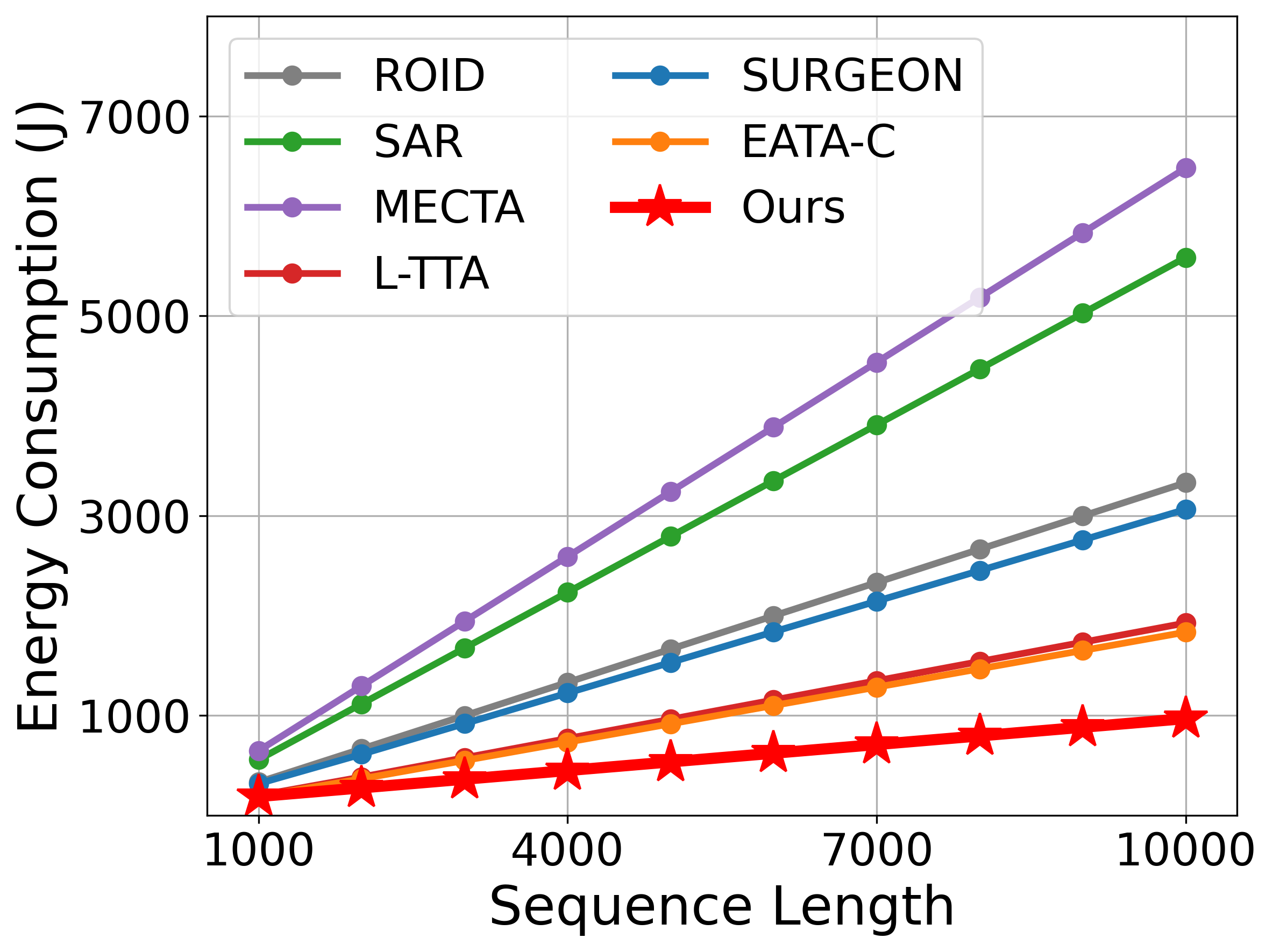}
  \label{fig:energy_consumption_bs1}} \hspace{0.1in}\subfigure[]{
		\includegraphics[width = 0.21\textwidth]{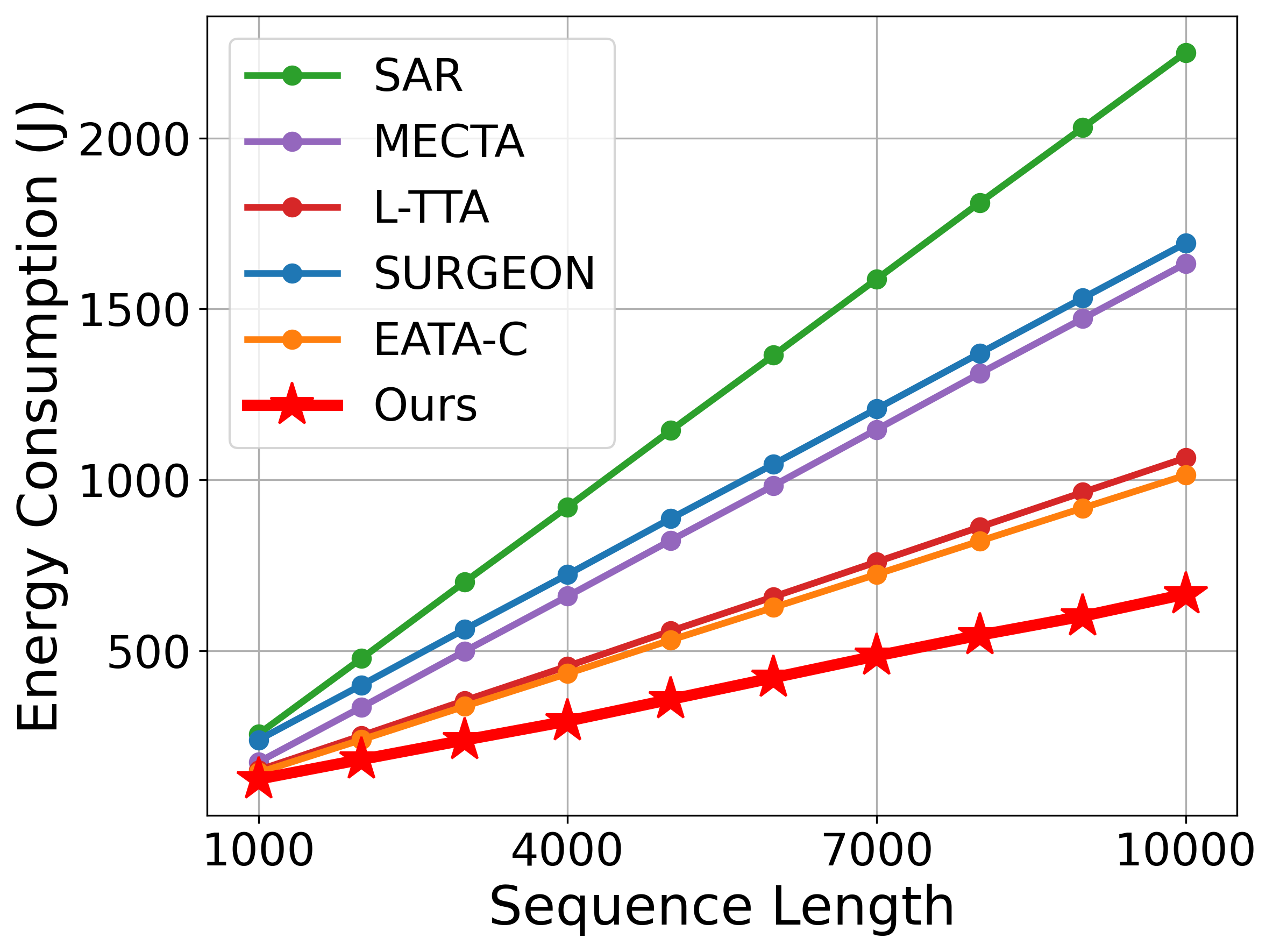}	
		\label{fig:energy_consumption_bs16}}
\vspace{-0.15in}
  \caption{Energy consumption for processing domain data sequences of varying lengths under batch size = (a) 1 and (b) 16.}
  \label{fig:energy_consumption}
\vspace{-0.1in}
\end{figure}

As shown in \Cref{tab:latency}, \sysname processes 10,000 samples in 239~s, 6.7$\times$ faster than MECTA (latency baseline). The speedup grows with sequence length because adaptation fires only on detected shifts, eliminating the per-batch overhead of C-TTA. \Cref{fig:energy_consumption} shows that \sysname saves 47.2\% energy over EATA-C at 10,000 frames, with consistent gains at batch size 16.\footnote{ROID cannot run at batch size 16 on device due to memory constraints.} For real-world shifts (e.g., weather, day/night transitions) that persist for hours, the savings are even larger.

In summary, \sysname outperforms nearly all baselines across metrics and offers the strongest trade-off among accuracy, memory footprint, and energy consumption, making TTA a practical, deployable solution for embodied devices.

\vspace{-2mm}
\subsection{Per-component Evaluation}

\begin{figure*}[t]
\centering
  \includegraphics[width = 0.88 \textwidth]{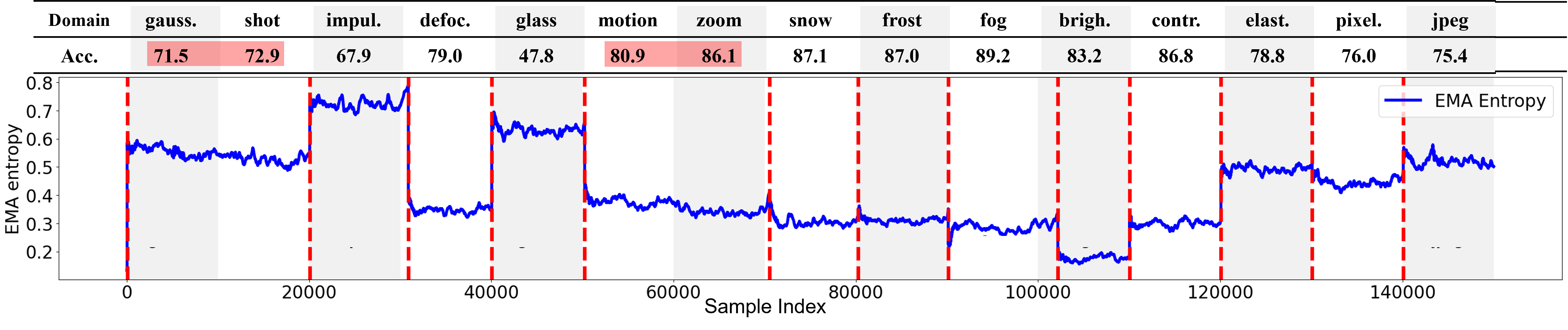}
  \vspace{-0.3cm}
  \caption{EMA entropy change along the data stream on CIFAR10-C. Domain transitions occur every 10,000 samples and are highlighted in alternating background (white-gray) color. Red dotted lines indicate detected shifts. The table above shows the corresponding accuracy on each domain.}
  \label{fig:detection_sequence}
  \vspace{-0.1in}
\end{figure*}

\subsubsection{Analysis of domain shift detection}
\label{sec: ablation_detection}

\Cref{fig:detection_sequence} illustrates the EMA entropy over CIFAR10-C streams, with detected adaptation triggers marked by red dotted lines. \sysname detects 13 of 15 domain shifts; the two missed transitions (Gaussian Noise~$\rightarrow$~Shot Noise, Motion Blur~$\rightarrow$~Zoom Blur) caused no accuracy degradation—performance actually improved by 1.4\% and 5.2\% respectively, confirming that skipping adaptation is the correct behavior when the shift is benign. For detected shifts, 7 of 13 are identified within 100 frames. To evaluate robustness to domain ordering, we test 10 randomly shuffled domain sequences: the detector identifies 13.9 of 15 shifts on average, with only 0.1 false triggers per sequence, while TTA accuracy remains stable at 79.5\%.

\begin{table}[t]
\caption{\textcolor{black}{Impact of different domain shift orders (S1--S10). Each sequence is a random shuffle of the 15 domains.}}
\label{tab:domain_sequence}
\centering
\resizebox{1.0\linewidth}{!}{
\begin{tabular}{@{}c|ccccccccccc@{}}
\toprule
\textbf{Sequences} & S1 & S2 & S3 & S4 & S5 & S6 & S7 & S8 & S9 & S10 & \textbf{Mean $\pm$ Std.} \\ 
\midrule
Triggered 
& 15 & 13 & 15 & 14 & 13 & 14 & 15 & 13 & 13 & \multicolumn{1}{c|}{14} 
& \textbf{13.9 $\pm$ 0.8} \\

Missed 
& 1 & 2 & 0 & 1 & 2 & 1 & 0 & 2 & 2 & \multicolumn{1}{c|}{1} 
& \textbf{1.2 $\pm$ 0.7} \\

False 
& 1 & 0 & 0 & 0 & 0 & 0 & 0 & 0 & 0 & \multicolumn{1}{c|}{0} 
& \textbf{0.1 $\pm$ 0.3} \\

Delay 
& 301 & 569 & 262 & 837 & 322 & 279 & 477 & 531 & 488 & \multicolumn{1}{c|}{523} 
& \textbf{459 $\pm$ 167} \\

Avg. acc. 
& 79.9 & 79.3 & 80.5 & 79.5 & 79.5 & 80.2 & 80.2 & 78.7 & 77.7 & \multicolumn{1}{c|}{79.7} 
& \textbf{79.5 $\pm$ 0.8} \\ 
\bottomrule
\end{tabular}}
\vspace{-0.1in}
\end{table}

\textcolor{black}{To further evaluate the robustness of the proposed shift detector to different domain orders, we construct 10 randomized domain sequences on CIFAR10-C. Specifically, we randomly shuffle the 15 corruption domains to generate 10 sequences (S1--S10). Each sequence contains 150,000 test samples, with domain transitions occurring every 10,000 samples, and is evaluated under the same batch-size-1 setting as the main experiments. \Cref{tab:domain_sequence} reports the number of detected shifts, missed shifts, false triggers, detection delay, and the resulting TTA accuracy. Across the 10 randomized sequences, the detector produces an average of 13.9 detection triggers, with only 1.2 missed shifts and 0.1 false triggers per sequence. The average detection delay is 459 samples. Moreover, the resulting TTA accuracy remains highly stable across different domain orders, averaging 79.5\% with a standard deviation of only 0.8\%. These results demonstrate that the proposed detector is robust to variations in domain ordering and that variations in detection timing do not substantially affect downstream adaptation performance.}

\begin{table}[t]
\centering

\begin{minipage}[t]{0.36\linewidth}
\centering
\caption{Threshold sensitivity on CIFAR10-C.}
\label{tab:threshold_sensitivity}
\resizebox{\linewidth}{!}{
\begin{tabular}{c|cc}
\toprule
$\mathbf{EMA_{\mathrm{thr}}}$ 
& Triggers 
& Accuracy (\%) \\
\midrule
0.03 & 19 & 80.1 \\
0.04 & 16 & 78.8 \\
0.05 & 15 & 78.5 \\
0.06 & 13 & 78.0 \\
0.07 & 12 & 75.5 \\
\bottomrule
\end{tabular}}
\end{minipage}
\hfill
\begin{minipage}[t]{0.63\linewidth}
\centering
\caption{On-demand adaptation ablation on CIFAR10-C.}
\label{tab:on_demand_ablation}
\resizebox{\linewidth}{!}{
\begin{tabular}{c|ccc}
\toprule
Method 
& Continual 
& Fixed-budget 
& On-demand \\
\midrule
Avg. Acc. (\%) 
& \textbf{78.2} 
& 77.8 
& 77.0 \\

Adapt. Samples
& 150K
& 1.92K
& \textbf{1.66K} \\

Workload Red. (\%)
& -- 
& 98.7 
& \textbf{98.9} \\
\bottomrule
\end{tabular}}
\end{minipage}

\end{table}





\textcolor{black}{
\textit{Threshold sensitivity.}
We further evaluate the sensitivity of the shift detector to different detection thresholds on CIFAR10-C. As shown in Table~\ref{tab:threshold_sensitivity}, decreasing $EMA_{\mathrm{thr}}$ from 0.07 to 0.03 progressively increases the number of adaptation triggers from 12 to 19, while the TTA accuracy increases from 75.5\% to 80.1\%. This trend reflects the expected accuracy--efficiency trade-off of on-demand adaptation: a smaller threshold makes the detector more sensitive and triggers adaptation more frequently, whereas a larger threshold avoids unnecessary adaptation but may tolerate more performance degradation before adaptation is activated. Notably, the accuracy remains relatively stable over a broad range of thresholds from 0.03 to 0.06, while the number of triggers decreases substantially. For CIFAR10-C, the standard deviation of the EMA entropy under the stable source distribution is $\sigma_E=0.012$, and the proposed setting $k=5$ corresponds to $EMA_{\mathrm{thr}}=0.06$, which is also the default threshold used in our experiments. These results demonstrate that the detector does not rely on a narrowly tuned threshold and support $k=5$ as a practical operating point that balances adaptation accuracy and frequency.}

\textcolor{black}{
\textit{Effect of on-demand adaptation.}
To isolate the contribution of the on-demand mechanism from the other components of \sysname, we remove both source-domain selection and the decoupled BN update, and compare three adaptation strategies on CIFAR10-C. Continual adaptation updates the model throughout each target domain. Fixed-budget adaptation instead uses only the first 128 samples of each domain for adaptation, serving as a control that reduces the adaptation budget without using the proposed shift detector. On-demand adaptation uses the same per-update budget of 128 samples, but performs adaptation only when a domain shift is detected. As shown in Table~\ref{tab:on_demand_ablation}, continual, fixed-budget, and on-demand adaptation achieve average accuracies of 78.2\%, 77.8\%, and 77.0\%, respectively. The fixed-budget strategy already achieves accuracy close to continual adaptation while reducing the number of adaptation samples from 150K to 1.92K, showing that continuously updating the model provides only marginal additional benefit. The on-demand mechanism further reduces the adaptation workload to 1.66K samples by skipping unnecessary updates, corresponding to a 98.9\% reduction relative to continual adaptation. These results demonstrate that the proposed detector further improves adaptation efficiency beyond simply imposing a fixed adaptation budget, which is particularly beneficial for resource-constrained embodied systems.}

\begin{table}[]
\caption{Evaluation of source model selection on CIFAR10-C and ImageNet-C.}
\label{tab:model_selection}
\centering
\resizebox{0.99\linewidth}{!}{
\begin{tabular}{@{}c|cc|cc|cc@{}}
\toprule
\multirow{2}{*}{Adapt from.} & \multicolumn{2}{c|}{Batch Size = 1} & \multicolumn{2}{c|}{Batch Size = 16} & \multicolumn{2}{c}{Batch Size = 64} \\ \cmidrule(l){2-7} 
 & CIFAR10-C & ImageNet-C & CIFAR10-C & ImageNet-C & CIFAR10-C & ImageNet-C \\ \midrule
Prev-Domain & 79.5 & 27.8 & 84.4 & 31.4 & 86.0 & 33.4 \\ \cmidrule(r){1-1}
Initial-Pool & 81.0 & \textbf{34.6} & \textbf{85.7} & 36.7 & \textbf{86.1} & \textbf{39.1} \\ \cmidrule(r){1-1}
Progressive-Pool & \textbf{81.7} & 34.5 & 85.1 & \textbf{37.5} & 86.0 & 37.7 \\ \bottomrule
\end{tabular}}
\vspace{-3mm}
\end{table}

\subsubsection{Analysis of source domain selection}
\label{sec:ablation_selection}

We use BN statistics from the second BN layer as the domain representation for candidate selection, following the design choice justified in Section~\ref{sec: similar candidate selection}. \Cref{tab:model_selection} compares adapting from the previous domain (\textit{Prev-Domain}), a pool built from source data (\textit{Initial-Pool}), and one built progressively from runtime history (\textit{Progressive-Pool}). Domain selection consistently outperforms \textit{Prev-Domain}, yielding up to 6.8\% accuracy gains on ImageNet-C. \textit{Progressive-Pool} matches \textit{Initial-Pool} closely, confirming effectiveness even without prior source data access.

\textcolor{black}{It is worth noting that Progressive-Pool is not expected to
monotonically converge to Initial-Pool, since the two candidate pools are constructed from different domain sources. Initial-Pool captures the domain diversity available in the source dataset,
whereas Progressive-Pool reflects historical target domains encountered during deployment. Therefore, their relative
performance depends on which pool provides better coverage of the current target distribution. This is reflected in Table~\ref{tab:model_selection}:
Progressive-Pool performs better in some settings, while Initial-Pool is advantageous in others. The two strategies thus
provide complementary options for deployments with and without prior access to source data, and can also be combined when both sources of candidate domains are available.}

\subsubsection{Analysis of decoupled BN update}

To assess the impact of adaptation sample count, we disable the shift detection module and adapt the model using varying numbers of samples (64 to 2048) from each domain in ImageNet-C, averaging accuracy across all 15 domains. Accuracy steadily improves and plateaus beyond 1024 samples, indicating that the new domain distribution is fully captured at that point. The decoupled update also reduces memory substantially: by separating BN statistics (forward-pass only, large batch) from BN parameters (backpropagation, small batch), \sysname reduces peak memory by $\sim$66\% compared to MECTA (414~MB vs.\ 1231~MB) at comparable accuracy.

\subsection{Generalizability Across Model Architectures}

\begin{table}[]
\caption{Adaptation accuracy (\%) on ImageNet-C using MobileNetV2 and MobileViT.}
\centering
\label{tab:mobilevit}
\resizebox{0.95\linewidth}{!}{
\begin{tabular}{@{}c|ccc|ccc@{}}
\toprule
\multirow{2}{*}{Method} & \multicolumn{3}{c|}{MobileNetV2} & \multicolumn{3}{c}{MobileViT} \\ \cmidrule(l){2-7} 
 & BS = 1 & BS = 16 & BS = 64 & BS = 1 & BS = 16 & BS = 64 \\ \midrule
SAR & 0.1 & 21.4 & 25.7 & 0.1 & 30.6 & 32.9 \\
MECTA & 0.2 & 0.6 & 0.7 & 0.2 & 0.2 & 0.2 \\
ROID & 0.2 & 21.5 & 25.4 & 0.2 & 30.0 & 33.1 \\
SURGEON & 0.2 & 1.38 & 23.5 & 0.2 & 2.51 & 28.4 \\
EATA-C & 0.1 & 21.8 & 26.2 & 0.9 & 27.3 & 32.5 \\
Ours & \textbf{14.8} & \textbf{22.5} & \textbf{26.8} & \textbf{22.4} & \textbf{32.8} & \textbf{33.3} \\ \bottomrule
\end{tabular}}
\end{table}

\sysname consistently outperforms all baselines on MobileNetV2 and MobileViT (\Cref{tab:mobilevit}), and remains the only method effective at batch size 1 across both architectures, confirming that its gains generalize beyond ResNet-50.

\section{Discussion}

\noindent\textbf{Accessing source data.}
In the TTA setting, direct access to the source domain dataset is typically restricted during adaptation. However, it remains unclear whether accessing the source data prior to adaptation, for purposes such as pre-computation or model customization, is permissible. For example, methods like EATA-C and L-TTA require access to source data before adaptation begins. In contrast, \sysname provides greater flexibility. When prior access to source data is available, it can be used to construct the candidate pool. If such access is not permitted, \sysname can instead employ a progressive strategy that builds the pool during runtime without requiring any source data.

\noindent\textbf{Generalizing to other modalities.} Beyond vision, embodied systems are typically equipped with diverse sensors (e.g., IMU, temperature, pressure, and light sensors) that produce continuous time-series streams and can also experience distribution shifts due to changes in the environment and sensor drift. Although this paper focuses on visual perception, the proposed on-demand TTA paradigm is broadly applicable to these modalities.



\section{Related Work}

\subsection{Embodied Perception}

Embodied systems need to perceive and act in open-world environments over long time horizons, where sensing conditions inevitably evolve with time, location, and operating context (e.g., illumination, weather, and backgrounds). Such domain shift is widely recognized as a key challenge for embodied AI~\cite{kunze2018artificial, wulfmeier2017addressing, siva2022robot}. To facilitate research in this setting, several benchmarks capture realistic long-term distribution changes, including CORe50~\cite{pmlr-v78-lomonaco17a}, EPIC-KITCHENS~\cite{Damen2021PAMI}, and SHIFT~\cite{sun2022shift}, enabling systematic study of long-term robustness under evolving environments.

\subsection{Continual Test-Time Adaptation}

C-TTA~\cite{wang2024search, liang2025comprehensive,xiao2024beyond} mitigates domain shift by adapting model parameters at test time using unsupervised objectives. Benz et al.~\cite{benz2021revisiting} highlighted the critical role of batch normalization (BN) in enabling adaptation under distribution shifts. Building on this insight, Wang et al.~\cite{wang2020tent} proposed TENT, which updates BN parameters via entropy minimization. SAR~\cite{niu2023towards} replaces BN with Group Normalization (GN) to enhance stability. To improve robustness under streaming settings, SoTTA~\cite{gong2023sotta} maintains a sample-balanced cache for more stable adaptation. ROID~\cite{marsden2024roid} and CoTTA~\cite{wang2022continual} are more memory-intensive, as they generate multiple augmented views and use a mean-teacher strategy for adaptation. Recent works have also focused on improving TTA efficiency. MECTA~\cite{hong2023mecta} replaces BN with a custom normalization layer to reduce memory usage. L-TTA~\cite{shin2024tta} introduces lightweight stem layers at shallow stages and adapts only these stems to reduce memory overhead.  SURGEON~\cite{ma2025surgeon} minimizes memory overhead during backpropagation by updating only the most critical weights. EATA-C~\cite{tan2025uncertainty} leverages uncertainty estimates to filter unreliable samples, improving robustness during test-time adaptation. 

Our work differs from existing research by introducing an on-demand TTA paradigm and a suite of techniques that improve adaptation effectiveness and efficiency under resource-constrained embodied deployment.

\subsection{Domain Shift Detection} 
\label{sec:literature_detection}

Domain shift detection is an essential part of \sysname, which monitors the distribution shift in the data stream to trigger the adaptation. Existing detection methods mainly utilize the martingale and statistics to measure the domain change. Luo~\cite{luo2022martingale} and Vovk~\cite{vovk2003testing} employ an auxiliary neural network to predict the martingale for each sample, serving as an indicator of domain shifts. However, they are memory-intensive because the auxiliary network (a dynamic CNN) must be continuously updated. Guy et al.~\cite{bar2024window} calculates a generalization bound for the source domain using the source dataset and identifies domain shifts by checking whether test samples exceed this established boundary. While this approach is less demanding in terms of memory, it is data-intensive, requiring substantial training data. Recently, Chakrabarty~\cite{chakrabarty2023simple} and Niloy~\cite{niloy2024effective} proposed using the mean of features extracted from a batch of data to represent the domain of the batch and reset the model to the source when the domain gap is over the threshold to achieve reliable C-TTA. However, these feature-based methods rely heavily on large batch sizes, making them unsuitable for online data streams where data arrives sequentially and in smaller batches.


Our detection approach is both lightweight and effective, offering a significant advantage over other methods by being adaptable to any batch-size configuration.

\vspace{-2mm}
\section{Conclusion}

This paper proposes a novel concept called on-demand TTA for resource-constrained embodied systems, which triggers adaptation only when a domain shift is detected. We introduce \sysname, a framework designed to realize on-demand TTA for edge devices. \sysname comprises three key components: domain shift detection to monitor distribution shifts on the fly, source domain selection to optimize the efficacy of the source model for adaptation, and decoupled BN adaptation to update the model efficiently under limited memory constraints. The experimental results show that \sysname outperforms baselines across nearly all evaluation settings in accuracy, latency, and energy consumption while maintaining comparable memory overhead, which promotes its deployment on embodied devices in real-world scenarios.

\ifCLASSOPTIONcaptionsoff
  \newpage
\fi



%
\bibliographystyle{IEEEtran}
\bibliography{references}

%








\end{document}